\documentclass[10pt,conference]{IEEEtran}
\usepackage{amsfonts}
\IEEEoverridecommandlockouts

\ifCLASSINFOpdf
\else
\fi
\usepackage{epsfig}
\usepackage{graphicx}
\usepackage{psfig}
\usepackage{epsf}
\usepackage[cmex10]{amsmath}
\usepackage{booktabs}
\usepackage{fancyhdr}
\usepackage{slashbox}
\usepackage{caption}   
\usepackage{subcaption}
\usepackage{balance}
\usepackage{bm}
\usepackage{underscore}
\usepackage{subcaption}
\usepackage{graphicx}
\usepackage{svg}

\begin{document}
\title{Data-and-Semantic Dual-Driven Spectrum Map Construction for 6G Spectrum Management}
\author{Jiayu Liu$^{\S}$, Fuhui Zhou$^{\S}$, Xiaodong Liu$^\dagger$, Rui Ding$^{\S}$, Lu Yuan$^{\S}$, and Qihui Wu$^{\S}$ \\
$^{\S}$Nanjing University of Aeronautics and Astronautics, Nanjing, China,\\
$^\dagger$Nanchang University, Nanchang, China.\\

Email: \emph{liujiayu_06@nuaa.edu.cn, zhoufuhui@ieee.org, xiaodongliu@whu.edu.cn,} \\ 
\emph{rui_ding@nuaa.edu.cn, yuanluylyl@gmail.com, wuqihui@nuaa.edu.cn}

\thanks{This paper has been accepted for presentation at the IEEE Global Communications Conference (GLOBECOM), Cape Town, South Africa, December 2024. This work was supported in part by Jiangsu Province Frontier Leading Technology Basic Research Project under Grant BK20222013, in part by National Key R\&D Program of China under Grant 2023YFB2904500, and in part by the National Natural Science Foundation of China under Grant 62222107. The corresponding author is Fuhui Zhou.}
}

\maketitle
\begin{abstract}
Spectrum maps reflect the utilization and distribution of spectrum resources in the electromagnetic environment, serving as an effective approach to support spectrum management. However, the construction of spectrum maps in urban environments is challenging because of high-density connection and complex terrain. Moreover, the existing spectrum map construction methods are typically applied to a fixed frequency, which cannot cover the entire frequency band. To address the aforementioned challenges, a UNet-based data-and-semantic dual-driven method is proposed by introducing the semantic knowledge of binary city maps and binary sampling location maps to enhance the accuracy of spectrum map construction in complex urban environments with dense communications. Moreover, a joint frequency-space reasoning model is exploited to capture the correlation of spectrum data in terms of space and frequency, enabling the realization of complete spectrum map construction without sampling all frequencies of spectrum data. The simulation results demonstrate that the proposed method can infer the spectrum utilization status of missing frequencies and improve the completeness of the spectrum map construction. Furthermore, the accuracy of spectrum map construction achieved by the proposed data-and-semantic dual-driven method outperforms the benchmark schemes, especially in scenarios with low sampling density.
\end{abstract}
\IEEEpeerreviewmaketitle
\section{Introduction}
\IEEEPARstart{W}{I}{T}{H} the proliferation of diverse wireless devices and the escalating demands of various business applications, spectrum scarcity has emerged as a major challenge confronting the sixth-generation (6G) wireless communication networks \cite{C. -X. Wang,Y. Zeng}. Moreover, there is a call for the full development and optimization of communication technologies based on different frequency bands, which in turn intensifies the competition for spectrum resources \cite{G. Pan,F. Shen}. Spectrum map construction is an effective technology capable of characterizing the distribution and utilization status of spectrum resources in the areas of interest at different frequencies \cite{Y. S. Reddy}. It is expected to alleviate spectrum scarcity and play a vital role in the dynamic management of spectrum resources  \cite{H. Xia}. Therefore, it is vital to construct accurate spectrum maps in the complex electromagnetic environments \cite{F. Zhou}. 

Existing spectrum map construction is primarily achieved through two schemes, namely model-based schemes and data-based schemes. Specifically, the model-based schemes rely heavily on predetermined propagation models. It is assumed that the signal propagation characteristics in the electromagnetic environment are fixed and known. However, the generalization capabilities of the model-based spectrum map construction methods are limited due to the difficulty of applying a specific propagation model to various radio propagation environments. Moreover, the dynamic nature of the complex electromagnetic environment makes it impractical to describe signal propagation with fixed models. 

Data-based schemes directly utilize collected spectral data to estimate the spectrum map in areas without measurements. However, the resolution of spectrum maps is directly proportional to the deployment density of sampling nodes in the area, which will greatly increase the cost of data collection, particularly in expansive outdoor scenarios. In this case, various spatial interpolation methods are adopted to enhance the resolution with an acceptable cost, such as inverse distance weighted \cite{Z. El-friakh}, K-nearest neighbor \cite{Y. Zhang}, kriging \cite{P. Maiti}, and tensor completion. Moreover, the accuracy of the data-based spectrum map construction methods is limited due to the irregular distribution of sampling nodes in the target area. Furthermore, the transmitted signal inevitably undergoes multi-path propagation in the real electromagnetic environment, including reflection, scattering, refraction, and diffraction. Consequently, the received signal is a superposition of different path signals from multiple different transmitters. Therefore, the signal propagation characteristics and received signals in real environments are more complex. Therefore, it is difficult to capture complex spatial relationships in a wide area utilizing traditional spatial interpolation data construction methods, especially in scenarios with low sampling density.

Deep learning promotes the development of spectrum map construction by learning the propagation rules of signals directly from raw spectrum data in an end-to-end manner. It is proven to be superior in improving the accuracy of spectrum map construction \cite{r9}-\cite{S. Zhang}. For instance, a deep completion autoencoder network structure was proposed in \cite{Y. Teganya} to learn the underlying rules of signal reflection, diffraction, and shadow in radio map estimation. However, the construction accuracy at low sampling density is still limited. In this case, to further enhance the accuracy of spectrum map construction under low sampling density, the authors in \cite{R. Levie} and \cite{S. Zhang} introduced city maps to assist neural networks in spectrum map construction. Specifically, a novel network, namely RadioUNet, was proposed in \cite{R. Levie} to estimate propagation path loss from the transmitting point to any receiving point in an urban environment. Then, the shadow effect during signal propagation in the outdoor spectrum map construction was further considered in \cite{S. Zhang}, and a two-stage learning framework was proposed for radio map estimation based on the conditional generative adversarial network. However, the cost of storing comprehensive and accurate city maps is expensive, and there is an urgent need to extract effective information from the maps to reduce the storage costs \cite{Y. Zeng}. Moreover, it is worth noting that the city maps adopted in \cite{R. Levie} and \cite{S. Zhang} were simply applied to the network without effectively extracting relevant information, which limited the contribution of city maps to the construction of spectrum maps. 

Note that existing model-based and data-based spectrum map construction studies primarily focus on the spatial correlation of spectrum data. The authors in \cite{K. Li} investigated the correlations of spectrum data in both time and spatial domains. Then, a spatial-temporal reconstruction network was presented for real-time spectrum map construction. However, the correlation of spectrum data in the frequency domain was not considered in \cite{K. Li}. It is limited to analyzing only a specific frequency band and is unable to construct spectrum maps for the entire frequency band. In order to solve this problem, the author in \cite{C. Wang} considered the frequency correlation of spectrum data. However, the author in \cite{C. Wang} mainly concentrated on the large-scale path loss during signal propagation and neglected the impact of small-scale multipath in the complex urban environments. Therefore, the investigation of the spectrum map construction within the complete frequency range from limited sampling data is imperative.

In this paper, a novel UNet-based data-and-semantic dual-driven method (DSD-UNet) is proposed to improve the accuracy of spectrum maps in the complex urban environments with dense communications. Specifically, binary city maps and sampling location maps are introduced as semantic knowledge to extract spatial dimension information that affects signal propagation. This is beneficial to neural network reasoning about the intrinsic mechanism of signal propagation in the spatial domain. Moreover, to further capture the signal propagation law in the frequency dimension, a joint frequency-space three-dimensional spectrum map model is exploited. This model leverages the frequency and spatial correlation of available spectral data to address the challenge of constructing spectrum maps for missing frequencies, a limitation of existing methods. The simulation results demonstrate that the proposed method is capable of inferring spectrum maps at missing frequencies by utilizing spectrum data from sampled frequencies. Furthermore, the accuracy of spectrum map construction and the convergence speed of the network are effectively improved by the integration of semantic knowledge, particularly in scenarios with low sampling density.

The remainder of this paper is organized as follows. The problem of the considered spectrum map construction is formulated in Section II. Section III presents the proposed joint frequency-space three-dimensional spectrum map
model and DSD-UNet. In Section IV, simulation results are presented. Finally, the paper is concluded in Section V.

\section{Problem Formulation}
For spectrum map construction in an urban environment, a target area $\mathcal{A}$ is selected. Specifically, there are $N_{T}$ transmitters and $N_{R}$ sampling receivers distributed randomly in the target area $\mathcal{A}$. The number of sampling receivers is determined according to the sampling density. In the considered target scenario, $N_{T}$ transmitters simultaneously emit signals at a certain frequency, which can be selected from $K+1$ available frequencies. Note that the entire available signal transmission frequencies are denoted as the set $\mathcal{F} =\left \{ f_{0},f_{1},\cdots  ,f_{K}  \right \}$. However, due to the complex propagation environment in urban environments and the limited performance of the sampling receiver, certain frequency bands may remain unmonitored. Correspondingly, the set of the measured frequency band monitored by the sampling receiver can be formulated as $\mathcal{F}_{s}  =\left \{ f_{1},\cdots  ,f_{K}  \right \}$. In this context, $f_{0}$ is set as the target inference frequency. In other words, the target inference frequency means that the receiver cannot collect the spectrum data pertaining to the frequency $f_{0}$. It is worth noting that any frequency within $\mathcal{F}$ can be set as the target frequency. 
Therefore, the key to spectrum map construction is to construct spectrum maps that describe the usage of transmission frequency set $\mathcal{F}$ using the spectrum data from the sampled frequency set $\mathcal{F}_{s}$.

Spatial discretization is first performed to establish a spectrum map construction model. Specifically, the target area $\mathcal{A}$ with $W\times W$ $\mathrm{m} ^{2}$ is evenly divided into a grid of size $N\times N$. The grid division interval is denoted as $\Delta d=W/N$, and each grid has a size of $\left ( \bigtriangleup d \right ) ^{2}$. Therefore, the resolution of the spectrum map is dependent on the grid scale $N\times N$. Moreover, the grid coordinates within $\mathcal{A}$ can be represented as $I=\left ( i,j \right )$, where $i,j=0,1,\cdots ,N-1$. The actual geographical location coordinate $\mathbf{x} _{i,j}$ of the center at grid $I$ can be expressed as $\mathbf{x} _{i,j} =\left ( \left ( i+0.5 \right )\bigtriangleup d, \left ( j+0.5 \right )\bigtriangleup d \right )$. In this paper, the grid interval in the considered scenario is set to be small to improve the resolution of the spectrum map. Based on this setting, it is reasonable to assume that the signal strength within the same grid exhibits only fluctuates slightly. Hence, for simplicity, the signal strength within each grid remains constant.

For the signal transmission frequency $f_{k}$, the received signal strength at the grid $I$ is defined as $P\left ( f_{k}, \mathbf{x} _{i,j}  \right )$, where $k=1,2,\cdots,K$. Consequently, a two-dimensional tensor ${\mathbf{S} } _k  \in \mathbb{R} ^{N\times N}$ is exploited to represent the incomplete spectrum map at the frequency $f_{k}$ of the target area. If the grid $I$ contains sampling receiver, the value of ${\mathbf{S} } _{k,i,j}$ is given by $P\left ( f_{k}, \mathbf{x} _{i,j}  \right )$. Otherwise, the value of ${\mathbf{S} } _{k,i,j}$ is set to be a constant value of zero. Similarly, the two-dimensional tensor ${\mathbf{P} } _k  \in \mathbb{R} ^{N\times N}$ is used to represent the real spectrum map at frequency $f_{k}$ and a two-dimensional tensor ${\mathbf{E} } _k  \in \mathbb{R} ^{N\times N}$ is used to represent the estimated spectrum map at frequency $f_{k}$. In this paper, the main objective is to use the sampled spectrum data $\left \{ {\mathbf{S} } _{k} \right \} _{k=1}^{K}$ to construct complete spectrum maps $\left \{ {\mathbf{E} } _{k} \right \} _{k=0}^{K}$ such that it is as close as possible to the real spectrum map $\left \{ {\mathbf{P} } _{k} \right \} _{k=0}^{K}$.

\section{Proposed Accurate Spectrum Map Construction Scheme}
\subsection{Joint Frequency-Space Three-Dimensional Spectrum Map Model}
\begin{figure}[!t]
\centering
\includegraphics[width=3.6 in]{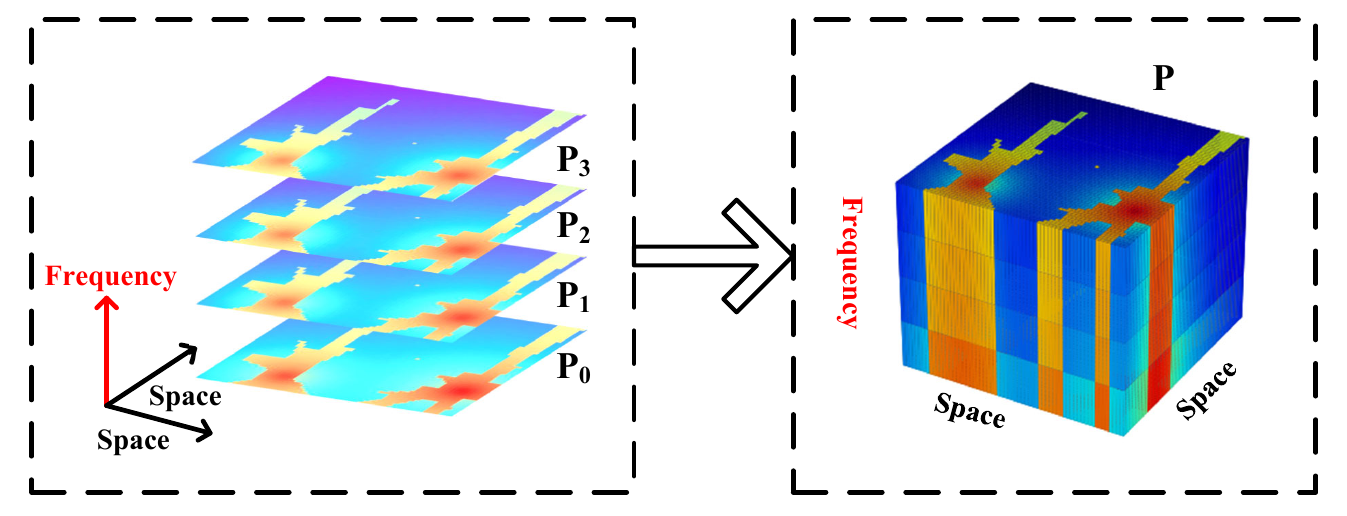}
\caption{Joint frequency-space three-dimensional spectrum map model.} \label{fig.1}
\end{figure}

Traditional spectrum map construction algorithms solely focus on the spatial domain correlation of spectrum data. It means that these construction algorithms cannot characterize the occupancy and utilization status of the undetected frequency bands in the sampling receiver. Therefore, a novel joint frequency-space three-dimensional spectrum map model is proposed by considering the correlation between spatial and frequency domains of spectrum data. As illustrated in Fig. \ref{fig.1}, the incomplete spectrum map ${\mathbf{S} } _k$ of the frequency $f_{k}$ at the target area is stacked along the frequency dimension in ascending order of frequency to build a two-dimensional stacked diagram with $K+1$ layers. Since the spectrum map of undetected frequency $f_{0}$ is the target to be inferred, it is set to blank. 

Therefore, the three-dimensional incomplete spectrum map model can be denoted as $\mathbf{S} \in {\mathbb{R} ^{N\times N\times \left ( K+1 \right ) }}$. Following the same procedure, a real three-dimensional spectrum map model denoted as $\mathbf{P} \in \mathbb{R} ^{N\times N\times \left ( K+1 \right ) }$, can be obtained by stacking the two-dimensional real spectrum map ${\mathbf{P} } _k$. Moreover, the estimated three-dimensional spectrum map model $\mathbf{E} \in \mathbb{R} ^{N\times N\times \left ( K+1 \right ) }$ can be obtained by stacking the two-dimensional estimated spectrum map ${\mathbf{E} } _k$. Moreover, it can be achieved by estimating ${\mathbf{S} }$ through an appropriate spectrum map construction method. The mathematical processing is given as
\begin{align}\label{27}\
\mathbf{E}=f\left (  \mathbf{S} \right ), 
\end{align}
where $f\left (  \cdot \right )$ represents the mapping rule of the DSD-UNet.

\begin{figure}[!t]
\centering
\includegraphics[width=3.0 in]{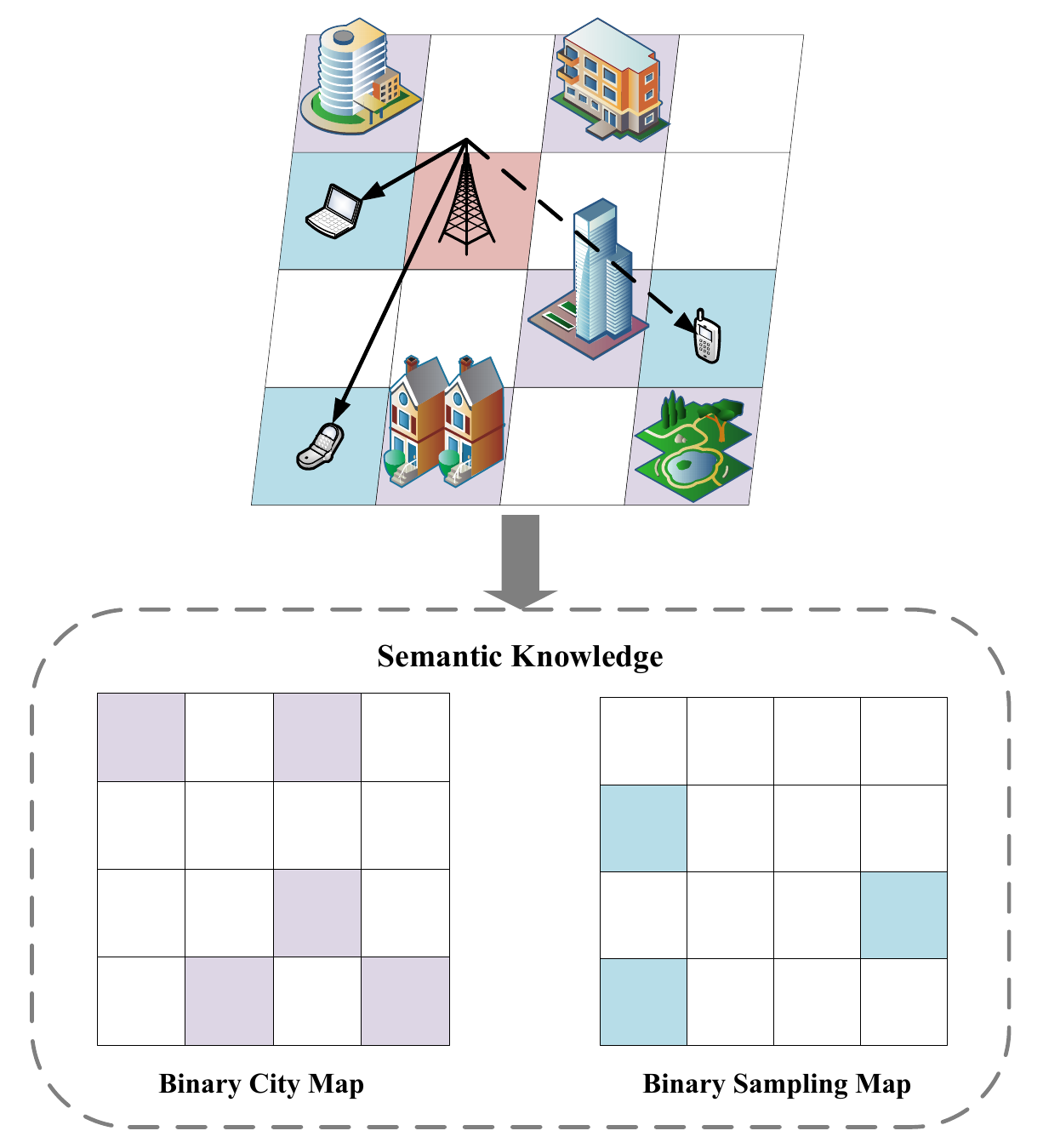}
\caption{Illustration of the semantic knowledge in the proposed method.} \label{fig.2}
\end{figure}
\begin{figure*}[!t]
\centering
\includegraphics[width=6.4 in]{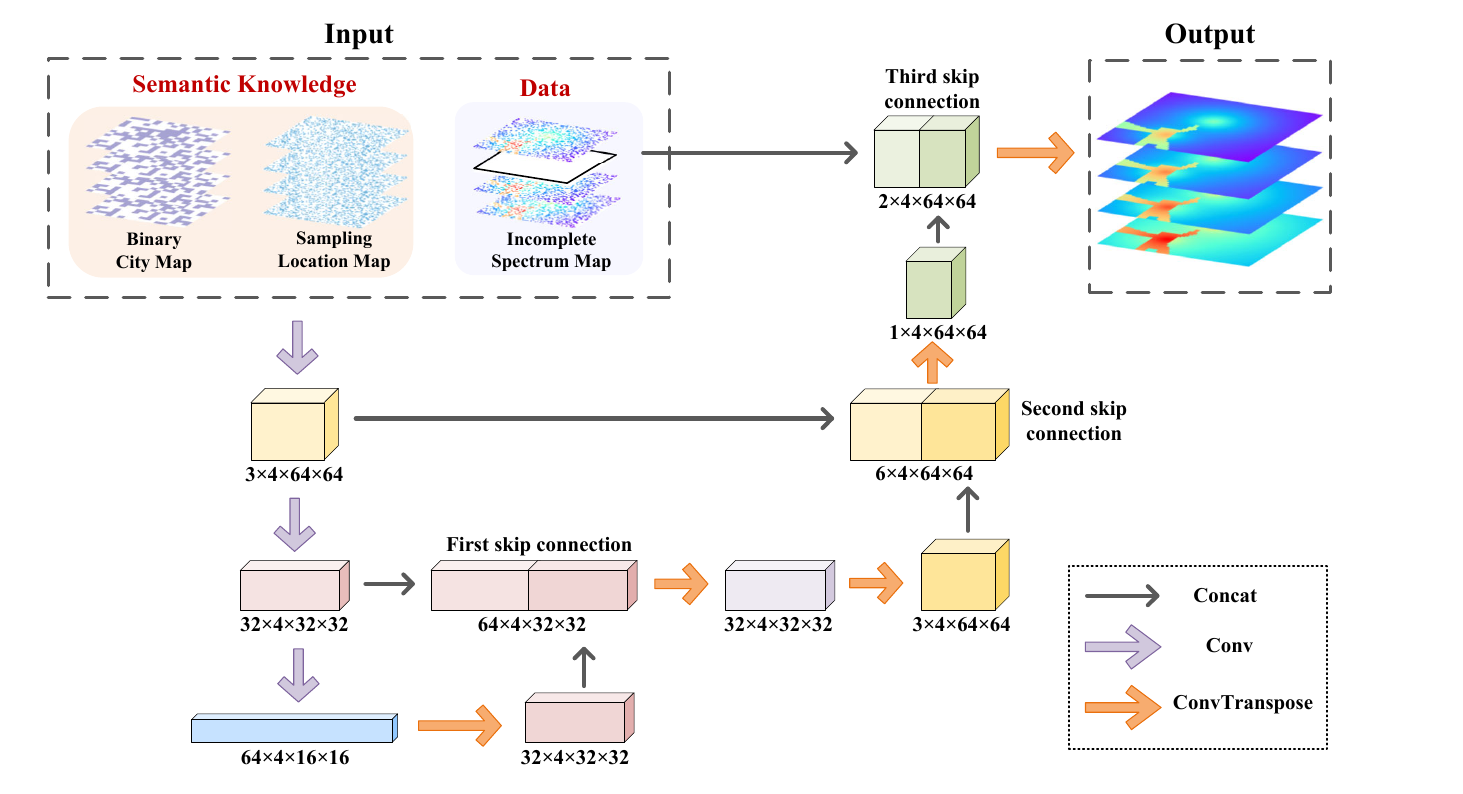}
\caption{The scheme of our proposed DSD-UNet.} \label{fig.3}
\end{figure*}

The DSD-UNet is trained to learn the rules of signal propagation in spatial and frequency domain. The goal of training is to minimize the error between the estimated spectrum map and the real spectrum map. The training goal is defined as
\begin{align}\label{27}\
\min \sum_{i=1}^{N} \sum_{j=1}^{N} \left \| \mathbf{E} -\mathbf{P}  \right \| _{F}^{2}, 
\end{align}
where $\left \| \cdot  \right \| _{F} $ represents the Frobenius norm of a tensor \cite{C. Wang}.

\subsection{Semantic Knowledge in Spectrum Map Construction}
In urban environments, signal propagation is typically blocked by buildings, which hinders the communication between the transmitter and the receiver. In this context, the signal undergoes multi-path propagation during transmission to the receiver, including reflection, refraction, scattering, and diffraction. Therefore, the signal propagation rules become more complicated, and the strength of the received signal is weakened. It can be seen that the propagation of signals through space is profoundly influenced by the location of buildings. Moreover, the distribution of sampling receivers also has an important impact on the accuracy of spectrum map construction. In the practical electromagnetic environment, sampling receivers are often distributed unevenly, which impacts the overall accuracy of the spectrum map construction. Fortunately, the distribution information of the sampled receiver locations facilitates a better understanding of signal propagation rules at different locations. 

Considering these facts, the semantic knowledge of binary city map and sampling location map is introduced to assist the proposed network in learning the propagation rules of signals. Specifically, a binary city map is defined as a two-dimensional tensor $\mathbf{Z} \in \mathbb{R} ^{N\times N}$. If there exists an obstacle obstructing signal propagation at a given grid point $I$, $\mathbf{Z}_{i,j}$ is assigned a value of 1. Conversely, if no obstacle hinders the grid point $I$, $\mathbf{Z}_{i,j}$ is assigned a value of 0. Based on this setting, a binary city map can be generated as,
\begin{align}\label{27}\
\mathbf{Z}_{i, j}=\left\{\begin{array}{ll}1, &  \mathrm{if} \  \mathrm{gird} \ I \ \mathrm{exists}  \  \mathrm{obstacles}  \\0, & \mathrm{else}, \end{array}\right.
\end{align}
where $\mathbf{Z}_{i,j}$ represents the value of binary city map $\mathbf{Z}$ at the grid point $I=\left ( i,j \right )$.

Similarly, a binary sampling location map is defined as a two-dimensional tensor $\mathbf{M}$. If the grid point $I$ contains one or more sampling receivers, the value is assigned to 1. If there is no sampling receiver in the grid point, the value is assigned to 0. Therefore, the sampling location semantics can be generated as,
\begin{align}\label{27}\
\mathbf{M}_{i, j}=\left\{\begin{array}{ll}1, & \mathrm{if} \  \mathrm{gird} \ I \ \mathrm{having}  \  \mathrm{sampling}  \  \mathrm{receivers}\\0, & \mathrm{else}, \end{array}\right.
\end{align}
where $\mathbf{M}_{i,j}$ represents the value of binary sampling location map $\mathbf{M}$ at the grid point $I=\left ( i,j \right )$.

\subsection{DSD-UNet for the Construction of Spectrum Maps}
The DSD-UNet structure is shown in Fig. \ref{fig.3}. Since the input of the construction model is three-dimensional frequency-space data, the two-dimensional semantics knowledge is transformed into three-dimensional semantic information by stacking two-dimensional binary city maps and sampling location maps of different frequencies along the frequency dimension. Firstly, the two-dimensional binary city map and the sampling location map at different frequencies are stacked along the frequency dimension.

Then, the binary city map, binary sampling location map, and the incomplete frequency-space three-dimensional spectrum map model are input into the proposed DSD-UNet in parallel. The convolution (Conv) operations are exploited to extract the features of the data, and thus, the compressed features are obtained. Moreover, the transposed convolution (ConvTranspose) operation is employed to restore the feature size to its original state while retaining the compressed features. In order to fully integrate the features derived from both the Conv and ConvTranspose operations, the features obtained by the Conv are skip-connected to the feature map obtained by the ConvTranspose. This process allows for a more comprehensive integration of the data within the network. Among them, in the third skip connection of DSD-UNet, the incomplete frequency-space three-dimensional spectrum map is duplicated and connected with the feature map obtained by the ConvTranspose to further utilize the collected spectrum data. This helps to enrich the neural network effectively with additional details.

Finally, the estimated frequency-space three-dimensional spectrum map can be obtained. It can be seen that the spectrum map of the missing target frequency is inferred and completed. Therefore, the incomplete spectrum map of the target area is constructed into a complete spectrum map. This approach demonstrates its capability to not only estimate incomplete spectrum maps at sampled frequencies but also generate a spectrum map at the missing frequency through spectrum data of the sampled frequency set.

\begin{figure*}
    \centering
    \begin{minipage}{1\textwidth}
        \begin{subfigure}{\textwidth}
            \includegraphics[width=7.2 in]{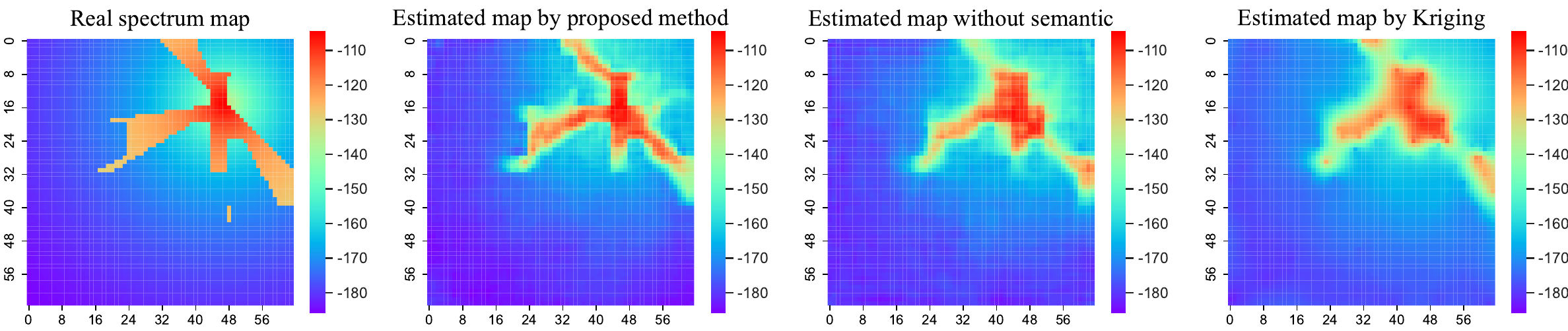}
            \caption{Single transmitter, sampling density = 5\%.}
            \label{fig:sub1}
        \end{subfigure}
        \begin{subfigure}{\textwidth}
            \includegraphics[width=7.2 in]{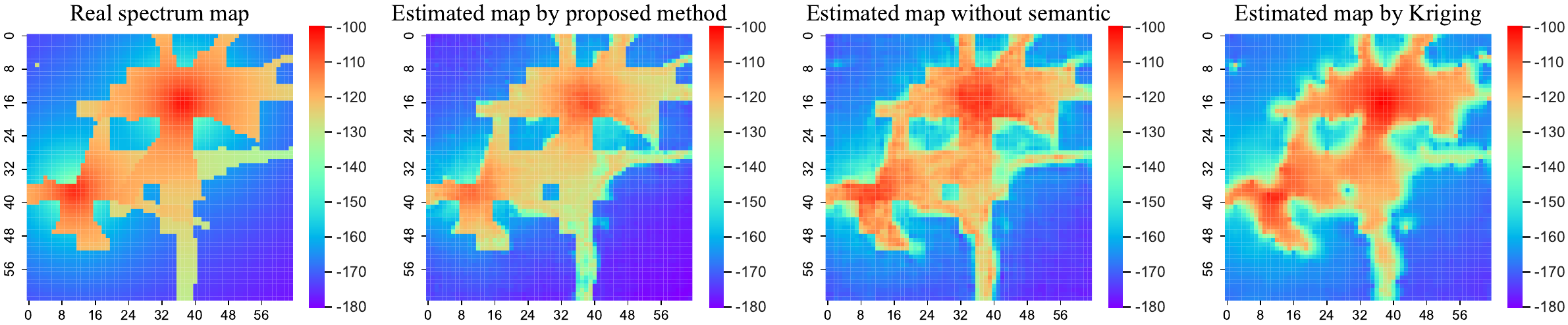}
            \caption{Multiple transmitter, sampling density = 20\%.}
            \label{fig:sub2}
        \end{subfigure}
    \end{minipage}
    \caption{The visualization results of the estimated spectrum maps under different methods.}
    \label{fig.5}
\end{figure*}

\section{Simulation Results}
In this section, the simulation results are given to compare the proposed DSD-UNet with the UNet-based scheme without semantics and the traditional kriging method. Subsequently, the efficacy and superiority of the proposed method are demonstrated.

\subsection{Parameter Setting}
In the simulation, the target area $\mathcal{A}$ of size $W\times W=256\times 256 \ \mathrm{m} ^{2} $ is divided into $N\times N=64\times 64$, with a grid interval of $4 \ \mathrm{m}$. Moreover, the entire signal transmission frequency is established at $\mathcal{F}=\left \{ 900,1500,1800,2100 \right \}  $ MHz, and $f_{0}=1800\mathrm{MHz}  $ is chosen as the target frequency. To enhance the convincing power of the results, the location, number, and transmit power are randomly set. Moreover, the sampling receivers are randomly deployed in the target area according to the sampling density of 5\%, 20\%, 35\%, and 50\%. Furthermore, the training set has 20,000 samples, and the test set has 1,000 samples for each sampling density. The training hyper-parameters of the proposed network are set as 30 epochs, a batch size of 4, and an initial learning rate of 0.0003. The two baseline comparison methods are the same network architecture but without semantic integration and the traditional kriging method, respectively.

\subsection{Performance Evaluation}

The visualization performance of the proposed DSD-UNet under different sampling densities is evaluated. Fig. \ref{fig.5} shows a comparison results of the visual effects of spectrum map construction among the proposed method and pure data-driven methods, as well as kriging methods. The results intuitively depict the performance of different methods in constructing spectrum maps for various scenarios with transmitter and sampling density settings. Specifically, Fig. \ref{fig.5}(a) shows the spectrum map construction effects of different methods when the sampling density is 5\% for a single transmitter. Moreover, Fig. \ref{fig.5}(b) describes the spectrum map construction effects of different methods when the sampling density is 20\% for multiple transmitters. It can be seen that the spectrum map constructed by the proposed method can be closer to the real spectrum map, regardless of the number of transmitters and the sampling density of receivers. Moreover, the proposed method is still capable of estimating spectrum maps with better accuracy under scenarios of low sampling density that result in reduced availability of spectrum data.

\begin{figure}[t!]
\flushleft
\includegraphics[width=3.7 in]{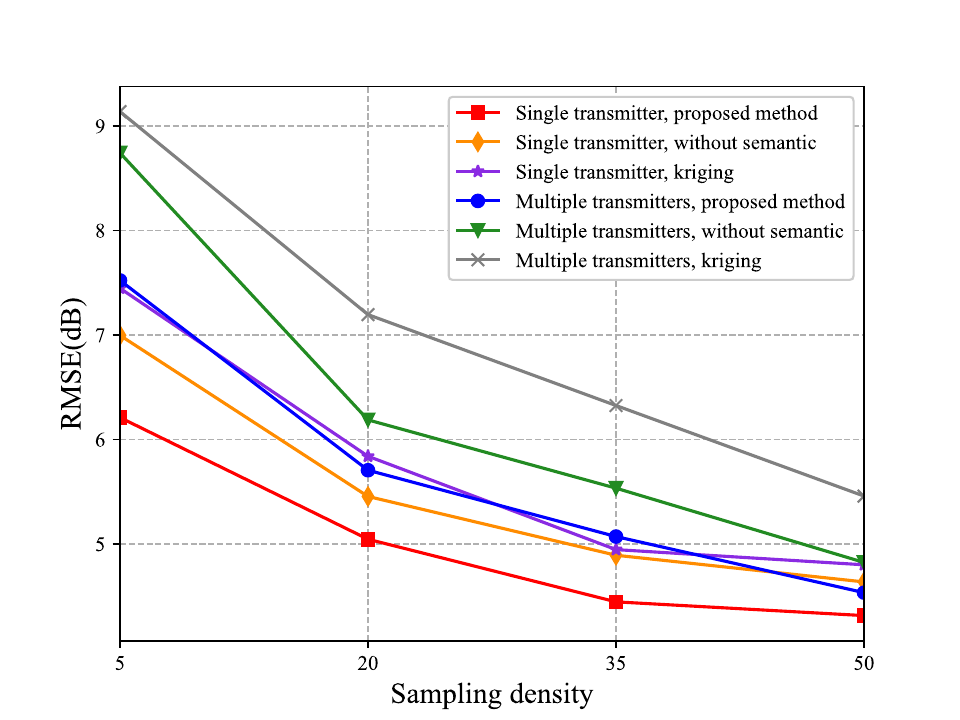}
\caption{Performance comparison under different sampling densities and transmitter scenarios.} \label{fig.6}
\end{figure}

In order to further describe the superiority of the proposed network, Fig. \ref{fig.6} shows the root-mean-square error (RMSE) performance of spectrum map construction under different sampling density and transmitter scenarios. The results show a decreasing trend in construction error as the sampling density increases. The reason is that a larger sampling density means that more spectrum data can be incorporated into these spectrum map construction methods to learn the propagation characteristics of the transmission signals. Moreover, it is seen that the accuracy of the proposed method outperforms the UNet-based method without semantics and kriging by approximately 10.72\% and 11.93\% in the single transmitter scenarios and multiple transmitters scenarios, respectively. Furthermore, it is also seen that the RMSE of the proposed method is 1dB lower than the UNet-based method without semantics and 1.43dB lower than the kriging method at low sampling density scenarios. These results effectively illustrate the proposed method under all sampling density and transmitter scenarios, especially at low sampling density and multiple transmitters scenarios.

\begin{figure}[!t]
\centering
\includegraphics[width=3.7 in]{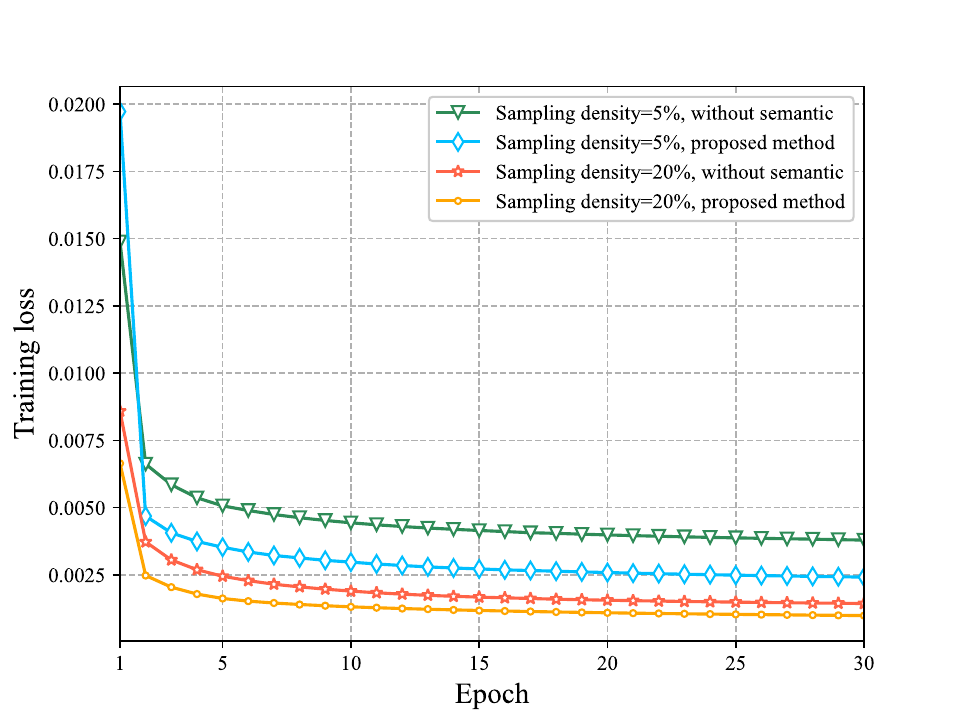}
\caption{Comparison of training loss under different sampling densities.} \label{fig.4}
\end{figure}

Fig. \ref{fig.4} illustrates the network convergence curves achieved by the proposed data-and-semantic dual-driven method and data-driven method without semantics under various sampling density scenarios. It can be seen that the proposed network just requires only 5 epochs to achieve convergence with the help of semantics knowledge. In contrast, the network without semantics knowledge takes 10 epochs to converge. The results demonstrate the effectiveness of incorporating semantics to accelerate the convergence of networks since semantic knowledge carries information that affects signal propagation, enabling the network to learn the signal distribution more rapidly.

\section{Conclusion}
A novel data-and-semantic dual-driven method was proposed for spectrum map construction in complex urban environments with dense communications. The information that affects signal propagation was fully exploited by introducing binary city maps and binary sampling location maps. Moreover, a joint frequency-space three-dimensional spectrum map model was established to realize the complete frequency set spectrum map construction. Simulation results showed that compared with the benchmark schemes, the proposed DSD-UNet can achieve better construction accuracy and faster convergence speed, especially in scenarios with low sampling density and multiple transmitters.

\end{document}